\documentclass[letterpaper,journal]{IEEEtran}

\usepackage{cite}
\usepackage{amsmath,amssymb}
\usepackage{graphicx}
\usepackage{booktabs,multirow,array,makecell}
\usepackage[caption=false,font=footnotesize,labelfont=sf,textfont=sf]{subfig}
\usepackage{xcolor}
\usepackage{url}
\usepackage[hidelinks]{hyperref}
\usepackage{stfloats}
\usepackage{adjustbox}

\newcommand{\tblhead}[1]{\thead{#1}}
\newcommand{\tblgroup}[2]{\multicolumn{#1}{c}{\textbf{#2}}}
\newcommand{\tblgroupbar}[2]{\multicolumn{#1}{c|}{\textbf{#2}}}

\title{I$^2$-KAN: Interpretable Intra- and Inter-Modal Multi-Pathway Representation Learning for Multimodal Sentiment Analysis}

\author{
Xiaoting Xie, Feng Qiao, Xianwei Zhang, Hailong Chen, Kaiwen Zheng, Joemon M. Jose, Xuri Ge$^{\dagger}$, Yujun Li$^{\dagger}$%
\thanks{
This work was supported by the Shandong Provincial Natural Science Foundation (ZR2026QC1054), the National Natural Science Foundation of China (62602376), and the Qingdao Science and Technology Benefiting the People Demonstration Program (25-1-5-cspz-21-nsh).

X. Xie, F. Qiao, X. Zhang, and H. Chen are with the School of Information Science and Engineering, Shandong University, Qingdao, Shandong, China (e-mail: 202412750@mail.sdu.edu.cn; 202420514@mail.sdu.edu.cn; 202512685@mail.sdu.edu.cn; 202332667@mail.sdu.edu.cn).

F. Qiao is also with the Department of Computer Science, University of Exeter, Exeter, UK.

K. Zheng and J. M. Jose are with the School of Computing Science, University of Glasgow, Glasgow, UK (e-mail: k.zheng.1@research.gla.ac.uk; joemon.jose@glasgow.ac.uk).

X. Ge is with the School of Artificial Intelligence, Shandong University, Jinan, Shandong, China (e-mail: xuri.ge@sdu.edu.cn).

Y. Li is with the School of Information Science and Engineering, Shandong University, Qingdao, Shandong, China (e-mail: liyujun@sdu.edu.cn).

$^{\dagger}$Corresponding authors: Xuri Ge and Yujun Li.
}}

\begin{document}

\maketitle

\begin{abstract}
Multimodal sentiment analysis (MSA) aims to infer sentiment by integrating linguistic, acoustic, and visual information. However, many fusion methods integrate modality-specific representations and cross-modal interactions into joint representations, making it difficult to determine whether predictive utility arises from a modality itself or from its interactions with others. More fundamentally, a modality's standalone predictive utility may not faithfully reflect its interaction-conditioned predictive utility. 
To address this issue, we propose I$^{2}$-KAN, an interpretable multi-pathway framework that explicitly maintains intra-modal and pairwise inter-modal information in separate structural pathways. Each pathway is refined by a Task-Decoupled Linear-Gated Denoising Bottleneck (TDL-GDB) to learn compact and task-relevant representations, followed by structured Kolmogorov--Arnold Network fusion that retains explicit pathway associations and supports multi-level interpretation. Extensive experiments on CMU-MOSI, CMU-MOSEI, and CH-SIMS v2 demonstrate strong overall performance. Ablation and interpretability analyses consistently show across the three benchmarks that low standalone predictive utility does not necessarily imply low interaction-conditioned predictive utility: a modality that appears weak in isolation can remain informative through pairwise interactions. These findings highlight the interaction-dependent nature of multimodal utility and suggest that modality utility should be characterized from both standalone and interaction-conditioned perspectives.
\end{abstract}


\section{Introduction}
\label{sec:introduction}

\IEEEPARstart{M}{ultimodal} sentiment analysis (MSA) aims to infer human affective states from heterogeneous behavioral signals, typically including language, acoustics, and vision. These modalities provide complementary affective cues, making their joint modeling important for understanding multimodal sentiment \cite{lai2023multimodal}. Owing to its relevance to human--computer interaction \cite{pantic2005affective}, social media understanding \cite{wei2023tackling}, conversational agents \cite{rashkin2019towards}, and human-centered artificial intelligence \cite{riedl2019human}, MSA has become an important problem in multimodal learning.

\begin{figure}[t]
    \centering
    \includegraphics[width=0.89\columnwidth]{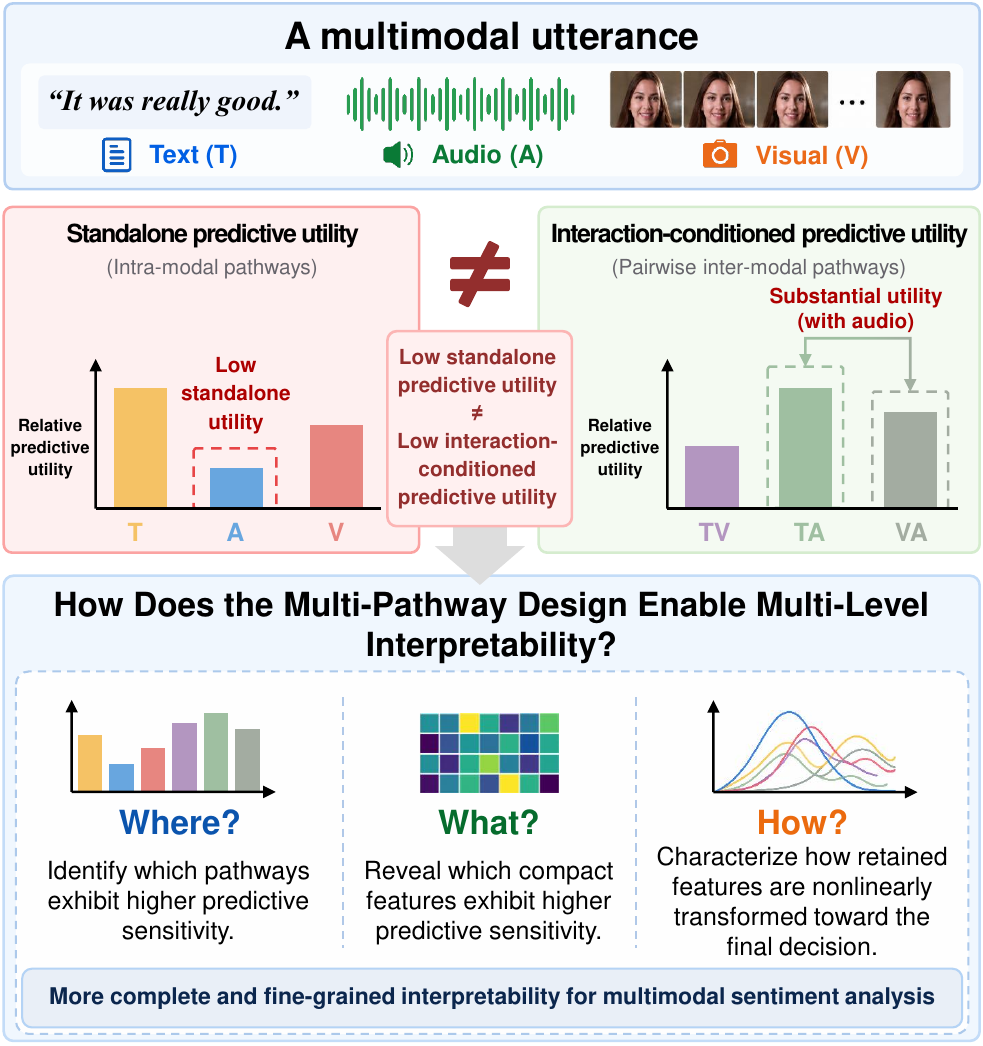}
    \caption{Motivation of I$^{2}$-KAN. Standalone predictive utility does not necessarily reflect interaction-conditioned predictive utility: a modality that exhibits limited utility in isolation can still provide substantial predictive utility through pairwise interactions. By explicitly maintaining intra-modal and pairwise inter-modal information within a multi-pathway structure, I$^{2}$-KAN supports three complementary levels of interpretation: where predictive sensitivity is concentrated across pathways, what compact features exhibit higher sensitivity within each pathway, and how those features are nonlinearly transformed toward the final decision.}
    \label{fig:motivation}
\end{figure}

A substantial body of research has investigated how to represent and fuse heterogeneous modalities. Early tensor-based methods explicitly modeled high-order multimodal interactions \cite{zadeh2017tensor,liu2018efficient}, while Transformer-based approaches captured long-range cross-modal dependencies through attention mechanisms \cite{tsai2019multimodal}. Later studies further explored modality-specific representation learning, auxiliary supervision, mutual-information maximization, and interaction-aware optimization \cite{hazarika2020misa,yu2021learning,han2021improving,luo2025triagedmsa,li2025mpid}. Despite this progress, cross-modal relations are often used as intermediate representations during prediction, making it difficult to separately trace the roles of standalone modalities and pairwise interactions within a unified decision structure.

A key issue is that a modality's standalone predictive utility does not necessarily reflect its utility when interacting with other modalities. As illustrated in Fig.~\ref{fig:motivation}, the acoustic modality may exhibit limited standalone predictive utility, whereas the Text--Audio and Visual--Audio pathways can exhibit substantially greater predictive utility in multimodal contexts. Thus, a modality that appears weak in isolation may still provide useful information through its interactions with other modalities. Evaluating only the original modalities may therefore provide an incomplete characterization of multimodal utility. Meanwhile, existing interpretability approaches mainly provide modality-level importance or analyze nonlinear fusion behavior \cite{tsai2020multimodal,luo2025towards}, making it difficult to jointly trace standalone and pairwise information sources throughout the decision process. More importantly, this discrepancy raises an interpretability requirement: a multimodal model should not only characterize the utility of different information sources, but also reveal where predictive sensitivity is concentrated, what compact features the prediction is sensitive to within each pathway, and how those features are transformed toward the final decision.

To address these issues, we propose I$^{2}$-KAN, an interpretable intra- and inter-modal multi-pathway framework that explicitly maintains standalone and pairwise inter-modal information as separate structural pathways. I$^{2}$-KAN consists of four stages. Modality-specific encoders first extract textual, acoustic, and visual representations, which are organized into a structured multi-pathway representation. A Task-Decoupled Linear-Gated Denoising Bottleneck (TDL-GDB) then independently refines each pathway through denoising compression, information-preserving shortcuts, and task-specific gated projections. Finally, the compact pathway representations are fused through a structured KAN \cite{liu2025kan}. By retaining explicit pathway associations throughout fusion, the framework supports pathway-level analysis of \emph{where} predictive sensitivity is concentrated, compressed-feature analysis of \emph{what} compact features the prediction is most sensitive to, and functional-edge analysis of \emph{how} those features are nonlinearly transformed toward the final prediction.

Extensive experiments on CMU-MOSI, CMU-MOSEI, and CH-SIMS v2 demonstrate competitive performance across different sentiment evaluation settings. Interaction-pathway and component-wise ablations validate the effectiveness of the multi-pathway representation, TDL-GDB, and KAN-based fusion. Multi-level analyses consistently show that low standalone predictive utility does not necessarily imply low interaction-conditioned predictive utility, supporting a more comprehensive characterization of modality utility beyond isolated modalities, while the structured multi-pathway design provides a traceable view of where, what, and how multimodal information participates in the learned decision process.

The main contributions can be summarized as follows:

\begin{itemize}
    \item We show across three benchmarks that standalone predictive utility and interaction-conditioned predictive utility can differ substantially. Motivated by this observation, we propose a multi-pathway framework (I$^{2}$-KAN) that separately maintains intra-modal and pairwise inter-modal information, enabling both forms of multimodal utility to be modeled and analyzed within a unified structure.

    \item We design a Task-Decoupled Linear-Gated Denoising Bottleneck (TDL-GDB) to learn compact and task-relevant representations for heterogeneous pathways. By combining denoising compression, an information-preserving shortcut, and task-specific gated projections, TDL-GDB allows different sentiment objectives to selectively preserve complementary information before subsequent global fusion.

    \item We construct a structured KAN fusion mechanism that retains explicit pathway associations throughout nonlinear fusion. Together with pathway-level, compressed-feature, and functional-edge analyses, this design provides a traceable view of how standalone and interaction-conditioned information participates in the final sentiment decision.
\end{itemize}

\section{Related Work}
\label{sec:related_work}

\subsection{Multimodal Fusion and Interaction Modeling}
\label{subsec:related_fusion}

Multimodal fusion is central to multimodal sentiment analysis because affective information is distributed across linguistic, acoustic, and visual signals. Early studies mainly focused on constructing expressive joint representations. Tensor Fusion Network (TFN) explicitly modeled unimodal, bimodal, and trimodal interactions through tensor products \cite{zadeh2017tensor}, Low-rank Multimodal Fusion (LMF) reduced the computational cost of tensor-based fusion through low-rank decomposition \cite{liu2018efficient}, and Recurrent Multistage Fusion Network (RMFN) decomposed cross-modal fusion into multiple recurrent stages \cite{liang2018multimodal}. MulT further employed directional cross-modal attention to effectively model long-range dependencies between unaligned multimodal sequences \cite{tsai2019multimodal}.

Subsequent studies increasingly emphasized modality-specific information and structured cross-modal relationships. MISA separated modality-invariant and modality-specific representations \cite{hazarika2020misa}, Self-MM introduced unimodal auxiliary supervision \cite{yu2021learning}, and MMIM retained task-relevant information through mutual-information maximization \cite{han2021improving}. MTAG modeled temporal and cross-modal relations using a heterogeneous attention graph \cite{yang2021mtag}. Related multimodal tasks have similarly explored explicit relational structures: Li et al. introduced dual-gated fusion for task-oriented intra- and inter-modal associations \cite{li2023dual}, developed an inductive framework for text-video representation alignment \cite{li2024multi}, and further modeled high-order intra- and inter-modal correlations through a variational multimodal hypergraph \cite{ijcai2025p907}.

More recent studies have explicitly investigated intra- and inter-modal relational learning in MSA. HGraph-CL models intra-modal structures using unimodal graphs and further captures inter-modal relations through hierarchical graph contrastive learning \cite{lin2022hgraph}. Semi-IIN independently captures intra- and inter-modal interactions through masked attention and dynamically regulates their importance using a gating mechanism under a semi-supervised learning framework \cite{lin2025semi}. TEMSA employs text-guided pairwise cross-modal mapping and reconstruction to jointly model inter-modal consistency and modality-specific information \cite{chen2025temsa}. More recently, I2RL progressively learns intra- and inter-modality relations through multi-scale intra-modal modeling, cross-modality shuffling, and reconstruction \cite{li2026i2rl}. These studies demonstrate the importance of explicitly considering different forms of multimodal relations rather than relying solely on holistic fusion.

Interaction-oriented analyses have also shown that strong multimodal performance does not necessarily imply effective cross-modal reasoning \cite{hessel2020does}. MRO decomposes predictions into unimodal, bimodal, and trimodal effects \cite{wortwein2022beyond}, while MPID \cite{li2025mpid}, TriagedMSA \cite{luo2025triagedmsa}, and CACR \cite{wang2026cacr} further explore modality-preserving, disagreement-aware, and interaction-aware representation learning. However, these approaches primarily focus on learning, selecting, or decomposing intra- and inter-modal relations during representation construction or prediction. In contrast, I$^{2}$-KAN explicitly maintains intra-modal and pairwise inter-modal representations as separate structural pathways throughout pathway-specific refinement and nonlinear fusion. This design preserves the source association of each compact representation and enables standalone modality behavior to be examined separately from pairwise interaction behavior at the pathway, compressed-feature, and functional-edge levels.

\subsection{Modality Imbalance and Robust Representation Learning}
\label{subsec:related_balance}

Different modalities often exhibit unequal information density, noise levels, and optimization difficulty, causing multimodal learning to be dominated by stronger modalities. Existing approaches address this issue through auxiliary supervision or optimization coordination. Self-MM constructs unimodal supervision signals from multimodal labels \cite{yu2021learning}, while MMPareto coordinates unimodal and multimodal gradient directions through Pareto optimization \cite{wei2024mmpareto}. More recent methods improve weak or under-optimized modalities through representation enhancement and reliability-aware optimization, including MFON \cite{zhang2025modal}, PaSE \cite{he2026pase}, and EBMC \cite{he2026enhance}.

Robust representation learning further seeks to suppress redundant or noisy multimodal information. The text-guided hierarchical noise eliminator t-HNE reduces intra- and inter-modal noise \cite{li2025t}, MISR learns stable sentiment representations under distribution shifts and cross-modal conflicts \cite{zhu2025misr}, and KAN-MCP combines a denoising information bottleneck with multimodal optimization to obtain compact unimodal representations for KAN fusion \cite{luo2025towards}. However, these methods primarily regulate modalities or their joint optimization and do not explicitly consider that standalone and interaction pathways may require different information preservation strategies. Moreover, different sentiment objectives may rely on different low-dimensional evidence. Our TDL-GDB therefore independently refines each intra-modal and interaction pathway while using task-decoupled gated projections to preserve task-relevant information before global fusion.

\subsection{Interpretability in Multimodal Sentiment Analysis}
\label{subsec:related_interpretability}

As multimodal fusion networks become increasingly complex, understanding how different information sources affect prediction has become important. Multimodal Routing provides global and instance-level modality explanations through routing weights \cite{tsai2020multimodal}, MTAG visualizes salient temporal and cross-modal graph connections \cite{yang2021mtag}, and MRO quantitatively decomposes unimodal and multimodal interaction effects \cite{wortwein2022beyond}. However, such importance scores or interaction decompositions do not directly reveal how individual compact features are nonlinearly transformed during fusion.

Kolmogorov--Arnold Networks (KANs) offer an intrinsically interpretable alternative by replacing scalar connection weights with learnable univariate functions on network edges \cite{liu2025kan}. KAN-MCP introduced KAN into MSA to visualize nonlinear mappings from compressed unimodal representations to sentiment predictions \cite{luo2025towards}, but its inputs remain limited to the three original modalities. In contrast, I$^{2}$-KAN preserves associations for both intra-modal and pairwise inter-modal pathways throughout KAN fusion, enabling interpretation at the pathway, compressed-feature, and functional-edge levels and allowing the roles of standalone and interaction-conditioned information to be traced toward the final prediction.

\section{Proposed Method}
\label{sec:method}

\begin{figure*}[!t]
    \centering
    \includegraphics[width=\textwidth]{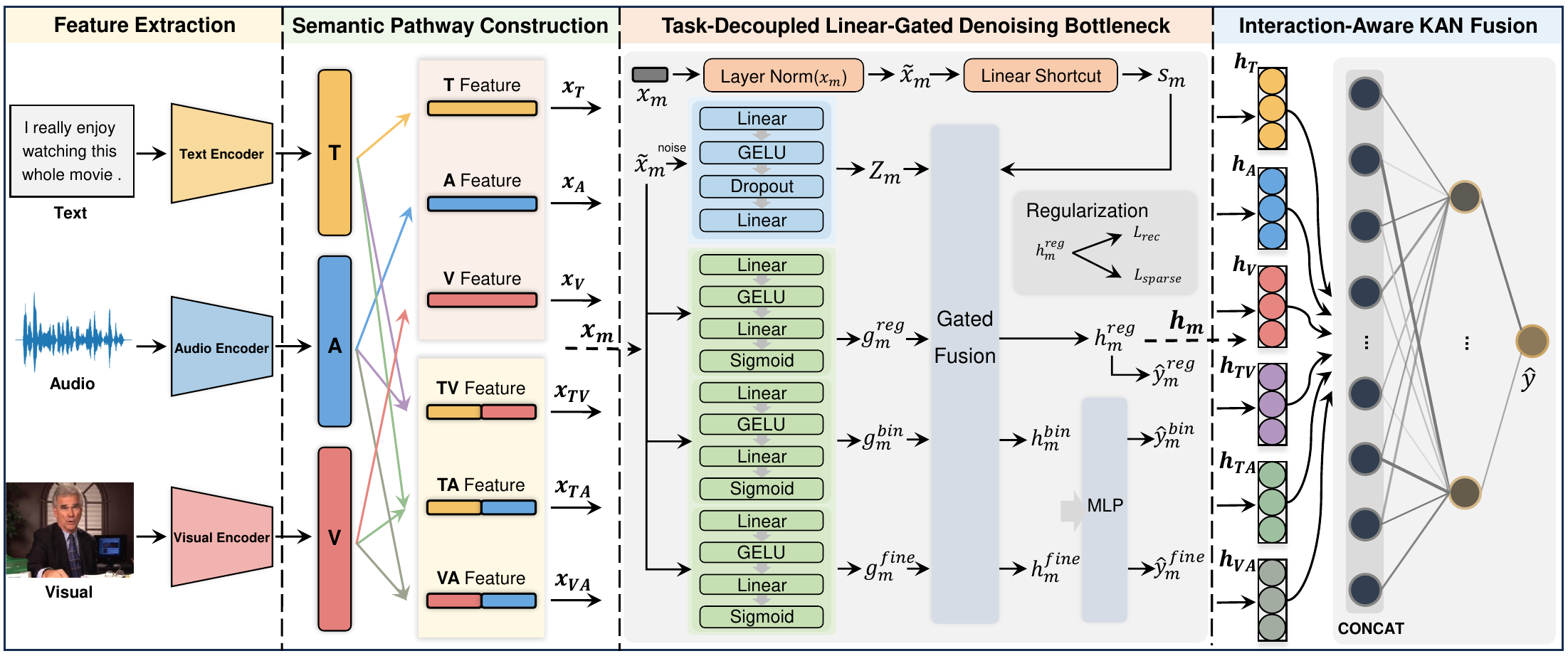}
    \caption{Overview of the proposed I$^{2}$-KAN framework. Modality-specific features are first organized into three intra-modal and three pairwise inter-modal semantic pathways. TDL-GDB independently refines each pathway into compact task-adaptive representations, which are subsequently integrated by the Interaction-Aware KAN Fusion module for sentiment prediction and multi-level interpretation.}
    \label{fig:framework}
\end{figure*}

In this section, we present I$^{2}$-KAN, an interpretable intra- and inter-modal multi-pathway framework for multimodal sentiment analysis, as illustrated in Fig.~\ref{fig:framework}. For notational simplicity, the Text, Audio, and Visual pathways are hereafter denoted as $T$, $A$, and $V$, respectively, while the Text--Visual, Text--Audio, and Visual--Audio interaction pathways are denoted as $TV$, $TA$, and $VA$. 

\subsection{Overall Framework}
\label{subsec:overall}

Given a multimodal utterance, I$^{2}$-KAN organizes multimodal affective evidence into source-traceable intra-modal and pairwise inter-modal pathways and processes them through four stages: Feature Extraction, Semantic Pathway Construction, Task-Decoupled Linear-Gated Denoising Bottleneck, and Interaction-Aware KAN Fusion. Modality-specific encoders first produce textual, acoustic, and visual representations, which are organized into three intra-modal and three pairwise inter-modal pathways. Each pathway is then independently refined by TDL-GDB into a compact task-adaptive representation. Finally, the six compact pathway representations are integrated by KAN in a fixed pathway order to produce the sentiment prediction while retaining explicit source-pathway associations for subsequent interpretation. The following subsections provide detailed formulations of these components.

\subsection{Feature Extraction and Semantic Pathway Construction}
\label{subsec:feature_representation}

Multimodal sentiment evidence can arise from both standalone modalities and their pairwise relationships. To preserve these two information sources explicitly, I$^{2}$-KAN organizes modality-specific representations into separate intra-modal and pairwise inter-modal pathways rather than immediately mixing them within a joint latent representation. Each pathway is subsequently refined and fused while retaining its structural source association.

Given a multimodal utterance, the textual, acoustic, and visual inputs are denoted as $X_t$, $X_a$, and $X_v$, respectively. Three modality-specific encoders are first employed to obtain the corresponding utterance-level representations:
\begin{equation}
z_t=E_t(X_t),\quad z_a=E_a(X_a),\quad z_v=E_v(X_v),
\end{equation}
where $E_t(\cdot)$, $E_a(\cdot)$, and $E_v(\cdot)$ denote the textual, acoustic, and visual feature extraction functions, respectively.

To explicitly preserve both standalone modality information and pairwise cross-modal information, the extracted representations are organized into six architecturally designated semantic pathways. The three intra-modal pathways are directly defined as
\begin{equation}
x_T=z_t,\quad x_A=z_a,\quad x_V=z_v.
\end{equation}
The three pairwise inter-modal pathways are constructed by concatenating the corresponding modality representations:
\begin{equation}
x_{TV}=[z_t;z_v],\quad x_{TA}=[z_t;z_a],\quad x_{VA}=[z_v;z_a],
\end{equation}
where $[\cdot\,;\cdot]$ denotes feature concatenation. These pairwise pathways do not assume that concatenation itself constitutes cross-modal interaction; rather, they preserve paired modality sources, from which pathway-specific nonlinear interactions are subsequently learned by TDL-GDB. No additional learnable interaction operator is introduced at this stage. Instead, each concatenated pairwise representation is subsequently processed by its own TDL-GDB, allowing modality-pair-specific information to be refined independently before global fusion.

The resulting six-pathway representation is therefore compactly written as
\begin{equation}
\mathcal{X}=\{x_T,x_A,x_V,x_{TV},x_{TA},x_{VA}\}.
\end{equation}
This construction preserves the structural source of each pathway throughout subsequent refinement and fusion, allowing standalone modality pathways and pairwise inter-modal pathways to remain distinguishable for task-specific modeling and subsequent interpretability analysis.

\subsection{Task-Decoupled Linear-Gated Denoising Bottleneck}
\label{subsec:tdlgdb}

Although the six semantic pathways preserve the structural origins of different affective information sources, their representations remain heterogeneous and may contain redundant or task-irrelevant components. Moreover, continuous sentiment regression, polarity discrimination, and fine-grained sentiment recognition may rely on different cues even within the same pathway. We therefore introduce the Task-Decoupled Linear-Gated Denoising Bottleneck (TDL-GDB), which combines perturbation-based denoising, nonlinear compression, an information-preserving shortcut, and task-specific gated projections. Each semantic pathway is equipped with an independently parameterized TDL-GDB, while the three task branches within the same pathway share the compressed candidates before task-specific selection.

For each semantic pathway $m\in\mathcal{M}$, where
\begin{equation}
\mathcal{M}=\{T,A,V,TV,TA,VA\},
\end{equation}
let $x_m\in\mathbb{R}^{d_m}$ denote its input representation. TDL-GDB first performs feature normalization,
\begin{equation}
\widetilde{x}_m=\operatorname{LN}(x_m),
\end{equation}
followed during training by Gaussian perturbation,
\begin{equation}
\widetilde{x}^{\,\mathrm{noise}}_m
=
\widetilde{x}_m+\epsilon_m,
\qquad
\epsilon_m\sim\mathcal{N}\!\left(\mathbf{0},\sigma^2\mathbf{I}_{d_m}\right),
\end{equation}
where $\operatorname{LN}(\cdot)$ denotes layer normalization, $\sigma$ controls the perturbation intensity, and $\mathbf{I}_{d_m}$ is the $d_m$-dimensional identity matrix. The perturbation is disabled during inference.

The perturbed representation is projected into a compact space through the nonlinear denoising encoder shown in Fig.~\ref{fig:framework}:
\begin{equation}
z_m
=
W^{e,2}_m
\mathcal{D}
\left[
\delta
\left(
W^{e,1}_m\widetilde{x}^{\,\mathrm{noise}}_m+b^{e,1}_m
\right)
\right]
+b^{e,2}_m,
\end{equation}
where $\delta(\cdot)$ and $\mathcal{D}(\cdot)$ denote GELU activation and dropout, respectively, and $z_m\in\mathbb{R}^{d_c}$ is the denoised compact representation. To reduce the risk of discarding discriminative cues during nonlinear compression, a parallel linear shortcut directly maps the normalized input into the same compact latent space:
\begin{equation}
s_m=W^s_m\widetilde{x}_m+b^s_m,
\end{equation}
where $s_m\in\mathbb{R}^{d_c}$. Hence, $z_m$ captures noise-robust nonlinear information, whereas $s_m$ provides a direct information-preserving projection.

TDL-GDB subsequently decouples the compact representation according to three sentiment objectives. Let $\mathcal{K}=\{\mathrm{reg},\mathrm{bin},\mathrm{fine}\}$ denote continuous regression, binary polarity classification, and fine-grained sentiment classification, respectively. For each task $k\in\mathcal{K}$, an independent two-layer gate is computed from the normalized pathway representation:
\begin{equation}
g_m^{k}
=
\sigma
\left[
W^{g,2}_{m,k}
\delta
\left(
W^{g,1}_{m,k}\widetilde{x}_m+b^{g,1}_{m,k}
\right)
+b^{g,2}_{m,k}
\right],
\end{equation}
where $\sigma(\cdot)$ denotes the sigmoid function and $g_m^{k}\in(0,1)^{d_c}$ controls the element-wise selection between the nonlinear and shortcut representations. The gated fusion is formulated as
\begin{equation}
u_m^{k}
=
g_m^{k}\odot z_m
+
\left(\mathbf{1}-g_m^{k}\right)\odot s_m,
\end{equation}
where $\odot$ denotes element-wise multiplication. Each task further employs an independent linear projection:
\begin{equation}
h_m^{k}
=
W^{k}_m u_m^{k}+b^{k}_m,
\qquad
k\in\mathcal{K}.
\end{equation}
Accordingly, $h_m^{\mathrm{reg}}$, $h_m^{\mathrm{bin}}$, and $h_m^{\mathrm{fine}}$ in Fig.~\ref{fig:framework} denote the regression-, polarity-, and fine-grained-specific representations, respectively.

During training, each task-specific representation is mapped to a pathway-level prediction through a lightweight MLP head:
\begin{equation}
\widehat{y}_m^{k}
=
P_m^{k}\!\left(h_m^{k}\right),
\qquad
k\in\mathcal{K}.
\end{equation}
Accordingly, $\widehat{y}_m^{\mathrm{reg}}$, $\widehat{y}_m^{\mathrm{bin}}$, and $\widehat{\mathbf{y}}_m^{\mathrm{fine}}$ denote continuous, binary, and fine-grained sentiment predictions, respectively. Their pathway-level training objectives are compactly written as
\begin{equation}
\begin{aligned}
\mathcal{L}_m^{\mathrm{reg}}
&=
\operatorname{SmoothL1}
\left(
\widehat{y}_m^{\mathrm{reg}},y
\right),\\
\mathcal{L}_m^{\mathrm{bin}}
&=
\operatorname{BCE}
\left(
\widehat{y}_m^{\mathrm{bin}},q^{\mathrm{bin}}
\right),\\
\mathcal{L}_m^{\mathrm{fine}}
&=
\operatorname{CE}
\left(
\widehat{\mathbf{y}}_m^{\mathrm{fine}},
\mathbf{q}^{\mathrm{fine}}
\right),
\end{aligned}
\end{equation}
where $y$, $q^{\mathrm{bin}}$, and $\mathbf{q}^{\mathrm{fine}}$ denote the corresponding supervision targets for the three tasks.

In addition to task-specific supervision, the regression representation is regularized through the reconstruction and sparsity constraints shown in Fig.~\ref{fig:framework}. A lightweight reconstructor recovers the normalized pathway representation as
\begin{equation}
\widehat{x}_m
=
R_m\!\left(h_m^{\mathrm{reg}}\right),
\end{equation}
and the two regularization terms are defined as
\begin{equation}
\mathcal{L}_m^{\mathrm{rec}}
=
\left\|
\widehat{x}_m-\widetilde{x}_m
\right\|_2^2,
\qquad
\mathcal{L}_m^{\mathrm{sparse}}
=
\left\|
h_m^{\mathrm{reg}}
\right\|_1.
\end{equation}

The pathway-level TDL-GDB objective is therefore
\begin{equation}
\begin{aligned}
\mathcal{L}_m^{\mathrm{TDL}}
={}&
\mathcal{L}_m^{\mathrm{reg}}
+\lambda_{\mathrm{bin}}\mathcal{L}_m^{\mathrm{bin}}
+\lambda_{\mathrm{fine}}\mathcal{L}_m^{\mathrm{fine}}\\
&+\lambda_{\mathrm{rec}}\mathcal{L}_m^{\mathrm{rec}}
+\lambda_{\mathrm{sparse}}\mathcal{L}_m^{\mathrm{sparse}},
\end{aligned}
\end{equation}
where the $\lambda$ coefficients control the relative strengths of the corresponding auxiliary objectives.

Only the regression-specialized representation $h_m^{\mathrm{reg}}$ is forwarded to the subsequent KAN fusion module, while the binary and fine-grained branches provide auxiliary supervision during training. For notational simplicity, we henceforth write
\begin{equation}
h_m\equiv h_m^{\mathrm{reg}}.
\end{equation}
The six compact representations entering KAN are therefore
\begin{equation}
\left\{
h_T,h_A,h_V,h_{TV},h_{TA},h_{VA}
\right\}.
\end{equation}

\subsection{Interaction-Aware KAN Fusion}
\label{subsec:kan_fusion}

After obtaining the refined representations of the six semantic pathways from TDL-GDB, I$^{2}$-KAN performs interaction-aware fusion through a structured Kolmogorov--Arnold Network (KAN). In conventional multimodal fusion methods, heterogeneous modality representations are commonly projected into a shared latent space through fusion operations, where modality-specific and cross-modal interaction information is implicitly mixed \cite{zadeh2017tensor,liu2018efficient,tsai2019multimodal}. Although these methods effectively capture multimodal dependencies, the structural origin of each modality and interaction pathway becomes difficult to trace during the fusion process. Therefore, I$^{2}$-KAN maintains explicit pathway assignments throughout nonlinear fusion, enabling a more transparent analysis of how different information sources participate in the final sentiment decision.

The six refined pathway representations generated by TDL-GDB are denoted as $h_T$, $h_A$, $h_V$, $h_{TV}$, $h_{TA}$, and $h_{VA}$, where each $h_m\in\mathbb{R}^{d_c}$ denotes the compact representation of the corresponding pathway. They are concatenated in a fixed pathway order to form
\begin{equation}
H=[h_T,h_A,h_V,h_{TV},h_{TA},h_{VA}],
\end{equation}
where $H\in\mathbb{R}^{6d_c}$. This structured formulation retains the source association of each compact representation, allowing intra-modal and pairwise inter-modal pathways to remain distinguishable during nonlinear fusion.

Unlike conventional MLPs with scalar connection weights, KAN associates each edge with a learnable univariate function $\phi_{ij}(\cdot)$~\cite{liu2025kan}. For two adjacent layers, the transformation is written as
\begin{equation}
y_j=\sum_i\phi_{ij}(x_i),
\end{equation}
where $x_i$ and $y_j$ denote the input and output neurons, respectively, and $\phi_{ij}(\cdot)$ represents the learnable function associated with the edge connecting them. The internal functional parameterization follows the adopted KAN implementation, with the corresponding configuration reported in Sec.~\ref{subsec:implementation_details}.

The final sentiment prediction is obtained as
\begin{equation}
\hat{y}=\operatorname{KAN}(H;\Theta_{\mathrm{KAN}}),
\end{equation}
where $\Theta_{\mathrm{KAN}}$ denotes the learnable parameters of KAN. During training, the final KAN prediction is supervised by the continuous sentiment label, while the pathway-level objectives defined in Eq.~(17) provide auxiliary supervision. Because each KAN input dimension retains an explicit association with its source pathway, the fusion process remains traceable from pathway-level inputs to learned nonlinear edge transformations. This structured representation supports subsequent pathway-level, compressed-feature, and functional-edge analyses at different levels of granularity, while the learned edge functions characterize how compact features are nonlinearly transformed toward the final prediction.

\section{Experiments}
\label{sec:experiments}

\begin{table*}[!t]
\centering
\caption{Comparison With Baselines on CMU-MOSI and CMU-MOSEI.}
\label{tab:main_mosi_mosei}
\renewcommand{\arraystretch}{1.09}
\setlength{\tabcolsep}{4.5pt}
\begin{tabular}{l|ccccc|ccccc}
\hline
\multirow{2}{*}{\tblhead{Model}} & \tblgroupbar{5}{CMU-MOSI} & \tblgroup{5}{CMU-MOSEI} \\
\cline{2-11}
& \tblhead{Acc7$\uparrow$} & \tblhead{Acc2$\uparrow$} & \tblhead{F1$\uparrow$} & \tblhead{MAE$\downarrow$} & \tblhead{Corr$\uparrow$} & \tblhead{Acc7$\uparrow$} & \tblhead{Acc2$\uparrow$} & \tblhead{F1$\uparrow$} & \tblhead{MAE$\downarrow$} & \tblhead{Corr$\uparrow$} \\
\hline
Graph-MFN\cite{zadeh2018multimodal} & 34.4 & 80.2 & 80.1 & 0.939 & 0.656 & 51.9 & 84.0 & 83.8 & 0.569 & 0.725 \\
MMIM\cite{han2021improving} & 45.0 & 85.1 & 85.0 & 0.738 & 0.781 & 53.1 & 85.1 & 85.0 & 0.547 & 0.752 \\
UniMSE\cite{hu2022unimse} & 48.7 & 86.9 & 86.4 & 0.691 & 0.809 & 54.4 & 87.5 & 87.5 & 0.523 & 0.773 \\
HGraph-CL\cite{lin2022hgraph} & -- & 86.2 & 86.2 & 0.717 & 0.799 & -- & 85.9 & 85.8 & 0.527 & 0.769 \\
HyCon\cite{mai2022hybrid} & 46.6 & 85.2 & 85.1 & 0.741 & 0.779 & 52.8 & 85.4 & 85.6 & 0.554 & 0.751 \\
ConFEDE\cite{yang2023confede} & 42.3 & 85.5 & 85.5 & 0.742 & 0.782 & \underline{54.9} & 85.8 & 85.8 & 0.522 & 0.780 \\
MGCL\cite{mai2023learning} & \underline{49.3} & 86.7 & 86.7 & 0.685 & 0.707 & 53.9 & 86.4 & 86.4 & 0.535 & 0.772 \\
MFM\cite{sun2024mfm} & 33.3 & 80.0 & 80.1 & 0.948 & 0.664 & 50.8 & 83.4 & 83.4 & 0.580 & 0.722 \\
ITHP\cite{xiao2024neuro} & 47.7 & 88.5 & 88.5 & 0.663 & 0.856 & 52.2 & 87.1 & 87.1 & 0.550 & \underline{0.792} \\
ULMD\cite{zhu2025multimodal} & 47.8 & 85.8 & 85.7 & 0.700 & 0.799 & 53.8 & 86.0 & 85.9 & 0.531 & 0.770 \\
MFON\cite{zhang2025modal} & 44.9 & 86.9 & 86.9 & 0.725 & 0.797 & 53.7 & 86.3 & 86.3 & 0.528 & 0.780 \\
KAN-MCP\cite{luo2025towards} & 48.3 & \underline{89.4} & \underline{89.4} & \underline{0.615} & \underline{0.857} & 53.9 & \underline{87.7} & \underline{87.6} & 0.522 & 0.788 \\
I2RL\cite{li2026i2rl} & 47.2 & 86.9 & 86.6 & 0.697 & 0.802 & 54.7 & 86.6 & 86.5 & 0.532 & -- \\
MDF\cite{jian2026mdf} & 48.3 & 85.8 & 85.9 & 0.692 & 0.810 & 54.8 & 86.6 & 86.4 & 0.525 & 0.788 \\
FUSE-Net\cite{yang2026factorize} & 49.2 & 85.8 & 85.9 & 0.688 & 0.798 & 54.3 & 85.8 & 86.1 & 0.527 & 0.772 \\
PSA-MF\cite{xie2026psa} & 46.5 & 86.4 & 86.2 & 0.686 & 0.807 & \textbf{55.0} & 86.3 & 86.3 & \underline{0.521} & 0.774 \\
\hline
I$^{2}$-KAN & \textbf{50.3} & \textbf{90.0} & \textbf{90.0} & \textbf{0.589} & \textbf{0.866} & 54.4 & \textbf{88.2} & \textbf{88.1} & \textbf{0.515} & \textbf{0.798} \\
\hline
\end{tabular}
\end{table*}

\begin{table}[!t]
\centering
\caption{Comparison With Baselines on the CH-SIMS v2 Dataset.}
\label{tab:main_simsv2}
\renewcommand{\arraystretch}{1.07}
\setlength{\tabcolsep}{5.4pt}
\begin{tabular}{l|ccccc}
\hline
\multirow{2}{*}{\tblhead{Model}} & \tblgroup{5}{CH-SIMS v2} \\
\cline{2-6}
& \tblhead{Acc3$\uparrow$} & \tblhead{Acc2$\uparrow$} & \tblhead{F1$\uparrow$} & \tblhead{MAE$\downarrow$} & \tblhead{Corr$\uparrow$} \\
\hline
EF-LSTM\cite{williams2018recognizing} & 73.5 & 80.1 & 80.0 & 0.309 & 0.700 \\
LF-DNN\cite{williams2018dnn} & 71.2 & 77.8 & 77.9 & 0.322 & 0.668 \\
TFN\cite{zadeh2017tensor} & 70.9 & 78.1 & 78.1 & 0.322 & 0.662 \\
LMF\cite{liu2018efficient} & 70.0 & 77.8 & 77.8 & 0.327 & 0.651 \\
MFN\cite{zadeh2018memory} & 72.7 & 79.4 & 79.4 & 0.301 & 0.712 \\
Graph-MFN\cite{zadeh2018multimodal} & 68.6 & 76.6 & 76.6 & 0.334 & 0.644 \\
MulT\cite{tsai2019multimodal} & 74.2 & 80.8 & 80.7 & 0.300 & 0.738 \\
MISA\cite{hazarika2020misa} & 68.9 & 78.2 & 78.3 & 0.342 & 0.671 \\
MAG-BERT\cite{rahman2020integrating} & 70.6 & 77.1 & 77.1 & 0.346 & 0.641 \\
Self-MM\cite{yu2021learning} & 72.7 & 78.7 & 78.6 & 0.315 & 0.691 \\
MMIM\cite{han2021improving} & 70.4 & 77.8 & 77.8 & 0.339 & 0.641 \\
AV-MC\cite{liu2022make} & 73.2 & 80.6 & 80.7 & 0.301 & 0.721 \\
KuDA\cite{feng2024knowledge} & \underline{74.3} & 80.2 & 80.1 & 0.289 & 0.741 \\
KAN-MCP\cite{luo2025towards} & \textbf{75.0} & 81.6 & 81.7 & \underline{0.281} & \underline{0.742} \\
FUSE-Net\cite{yang2026factorize}
& 73.7 & 79.6 & 79.5 & 0.297 & 0.712 \\
CmIR\cite{mai-han-2026-learning} & \textbf{75.0} & \underline{81.9} & \underline{82.0} & 0.286 & \underline{0.742} \\
\hline
I$^{2}$-KAN & 72.4 & \textbf{85.6} & \textbf{85.6} & \textbf{0.280} & \textbf{0.746} \\
\hline
\end{tabular}
\end{table}

\subsection{Datasets and Evaluation Metrics}
\label{subsec:datasets_metrics}

\textbf{CMU-MOSI~\cite{zadeh2016multimodal}.} CMU-MOSI is a widely used English multimodal sentiment analysis benchmark containing 2,199 opinion utterances extracted from 93 videos. Each utterance is annotated with a continuous sentiment score ranging from $-3$ to $+3$. Following the standard split, 1,284 utterances are used for training, 229 for validation, and 686 for testing.

\textbf{CMU-MOSEI~\cite{zadeh2018multimodal}.} CMU-MOSEI is a large-scale English multimodal sentiment dataset with sentiment annotations ranging from $-3$ to $+3$. Although the dataset also provides emotion labels, only sentiment annotations are considered in this work. We follow the standard partition containing 16,326 training, 1,871 validation, and 4,659 test utterances.

\textbf{CH-SIMS v2~\cite{liu2022make}.} CH-SIMS v2 is a Chinese multimodal sentiment dataset collected from diverse scenarios, including films, television programs, interviews, and talk shows. Each sample is annotated with a continuous sentiment score ranging from $-1$ to $1$. We use the official split containing 2,722 training, 647 validation, and 1,034 test samples.

For CMU-MOSI and CMU-MOSEI, we report seven-class accuracy (Acc7), binary accuracy (Acc2), weighted F1 score, mean absolute error (MAE), and Pearson correlation coefficient (Corr). Acc7 maps continuous sentiment predictions to seven levels from $-3$ to $+3$. Following the commonly adopted positive/negative evaluation protocol, samples with zero ground-truth sentiment are excluded when computing Acc2 and F1. For CH-SIMS v2, we report three-class accuracy (Acc3), Acc2, weighted F1, MAE, and Corr. Acc3 divides the sentiment space into negative, neutral, and positive classes using $-0.2$ and $0.2$ as thresholds. Higher Acc7/Acc3, Acc2, F1, and Corr and lower MAE indicate better performance.

\subsection{Implementation Details}
\label{subsec:implementation_details}

All experiments are implemented in PyTorch. For CMU-MOSI and CMU-MOSEI, DeBERTa-v3-base, COVAREP, and FaceNet512 are adopted to extract textual, acoustic, and visual representations, respectively. For CH-SIMS v2, Chinese BERT, OpenSMILE, and OpenFace are used for the three modalities. Temporal mean pooling is applied when necessary to obtain utterance-level representations. The compact dimension of each semantic pathway is set to 3 for all datasets. The intermediate dimension of TDL-GDB is set to 1024, 518, and 1024 for CMU-MOSI, CMU-MOSEI, and CH-SIMS v2, respectively, while the number of hidden nodes in KAN is set to 4, 4, and 3. The internal functional parameterization of KAN follows the default configuration of the adopted implementation~\cite{liu2025kan}.

All models are optimized using AdamW with a weight decay of $0.01$. For CMU-MOSI, the batch size is 8 and the model is trained for 35 epochs, with learning rates of $1\times10^{-5}$ for the text-related parameters and $1\times10^{-3}$ for the remaining parameters. For CMU-MOSEI, the batch size is 64 and the model is trained for 10 epochs, with learning rates of $2\times10^{-5}$ and $1\times10^{-2}$ for the text-related and remaining parameters, respectively. A linear learning-rate scheduler with a warm-up ratio of $0.5$ is applied to both datasets. For CH-SIMS v2, the batch size is 32 and the model is trained for 50 epochs using a learning rate of $1\times10^{-3}$ without an additional learning-rate scheduler. For auxiliary supervision, the fine-grained branch follows the seven-level sentiment targets on CMU-MOSI and CMU-MOSEI and the corresponding three-level targets on CH-SIMS v2, while the binary branch follows the corresponding positive--negative polarity definition. All reported results of I$^{2}$-KAN and its variants are averaged over five runs with different random seeds.

\begin{table*}[!t]
\centering
\caption{Analysis of Pairwise Interaction Pathways on CMU-MOSI and CMU-MOSEI. All configurations retain the intra-modal pathways $T$, $A$, and $V$.}
\label{tab:interaction_mosi_mosei}
\renewcommand{\arraystretch}{1.09}
\setlength{\tabcolsep}{4.5pt}
\begin{tabular}{l|ccccc|ccccc}
\hline
\multirow{2}{*}{\tblhead{Interactions}} & \tblgroupbar{5}{CMU-MOSI} & \tblgroup{5}{CMU-MOSEI} \\
\cline{2-11}
& \tblhead{Acc7$\uparrow$} & \tblhead{Acc2$\uparrow$} & \tblhead{F1$\uparrow$} & \tblhead{MAE$\downarrow$} & \tblhead{Corr$\uparrow$} & \tblhead{Acc7$\uparrow$} & \tblhead{Acc2$\uparrow$} & \tblhead{F1$\uparrow$} & \tblhead{MAE$\downarrow$} & \tblhead{Corr$\uparrow$} \\
\hline
None & 44.9 & 86.5 & 86.5 & 0.704 & 0.808 & 49.2 & 86.7 & 86.6 & 0.578 & 0.753 \\
VA & 47.4 & 86.2 & 86.2 & 0.699 & 0.811 & 52.6 & 87.7 & 87.7 & 0.539 & 0.763 \\
TV & 47.0 & 88.2 & 88.1 & 0.648 & 0.841 & 54.2 & 88.1 & 88.1 & 0.516 & 0.789 \\
TA & 47.8 & 88.7 & 88.6 & 0.631 & 0.856 & \textbf{54.8} & 87.1 & 87.0 & 0.523 & 0.784 \\
TV+VA & 48.1 & 89.4 & 89.4 & 0.628 & 0.856 & 54.2 & 87.7 & 87.7 & 0.521 & 0.787 \\
TV+TA & 49.1 & 89.0 & 88.9 & 0.620 & 0.855 & 54.1 & 87.4 & 87.4 & 0.519 & 0.787 \\
TA+VA & 48.3 & 88.7 & 88.6 & 0.627 & 0.853 & 53.0 & 87.5 & 87.3 & 0.525 & 0.785 \\
Full & \textbf{50.3} & \textbf{90.0} & \textbf{90.0} & \textbf{0.589} & \textbf{0.866} & 54.4 & \textbf{88.2} & \textbf{88.1} & \textbf{0.515} & \textbf{0.798} \\
\hline
\end{tabular}
\end{table*}

\begin{table}[!t]
\centering
\caption{Analysis of Pairwise Interaction Pathways on CH-SIMS v2.}
\label{tab:interaction_sims}
\renewcommand{\arraystretch}{1.07}
\setlength{\tabcolsep}{5.4pt}
\begin{tabular}{l|ccccc}
\hline
\multirow{2}{*}{\tblhead{Interactions}} & \tblgroup{5}{CH-SIMS v2} \\
\cline{2-6}
& \tblhead{Acc3$\uparrow$} & \tblhead{Acc2$\uparrow$} & \tblhead{F1$\uparrow$} & \tblhead{MAE$\downarrow$} & \tblhead{Corr$\uparrow$} \\
\hline
None & 66.1 & 81.4 & 81.4 & 0.316 & 0.703 \\
VA & 70.4 & 83.8 & 83.8 & 0.300 & 0.736 \\
TV & 71.0 & 84.6 & 84.6 & 0.298 & 0.734 \\
TA & 70.4 & 84.3 & 84.3 & 0.301 & 0.738 \\
TV+VA & 71.0 & 84.6 & 84.5 & 0.308 & 0.727 \\
TV+TA & 70.3 & 84.1 & 84.0 & 0.305 & 0.721 \\
TA+VA & 69.5 & 84.0 & 84.0 & 0.298 & 0.734 \\
Full & \textbf{72.4} & \textbf{85.6} & \textbf{85.6} & \textbf{0.280} & \textbf{0.746} \\
\hline
\end{tabular}
\end{table}

\subsection{Performance Comparison}
\label{subsec:performance_comparison}

\textbf{Baselines.} We compare I$^{2}$-KAN with representative and recent methods covering multimodal fusion, representation learning, modality optimization, and interaction modeling. The compared methods and their corresponding references are summarized in Tables~\ref{tab:main_mosi_mosei} and~\ref{tab:main_simsv2}. The best and second-best results are highlighted in bold and underlined, respectively.

As shown in Table~\ref{tab:main_mosi_mosei}, I$^{2}$-KAN achieves the best results across all five metrics on CMU-MOSI. In particular, it improves both classification-oriented metrics (Acc7, Acc2, and F1) and regression-oriented metrics (MAE and Corr), demonstrating consistent performance across discrete and continuous sentiment evaluation.

On CMU-MOSEI, I$^{2}$-KAN achieves the best Acc2, F1, MAE, and Corr while remaining competitive in Acc7. These results indicate that the proposed framework maintains strong performance on the substantially larger CMU-MOSEI benchmark across both classification and regression metrics.

As shown in Table~\ref{tab:main_simsv2}, I$^{2}$-KAN achieves the best results in Acc2, F1, MAE, and Corr on CH-SIMS v2. In particular, the improvements in Acc2 and F1 indicate a clear advantage in sentiment polarity discrimination. Overall, the results across the three benchmarks confirm the effectiveness of I$^{2}$-KAN under different languages and sentiment annotation settings.

\subsection{Analysis of Pairwise Interaction Pathways}
\label{subsec:interaction_analysis}

To examine the effect of explicit pairwise interaction pathways and how different pathway combinations affect sentiment prediction, we evaluate all possible combinations of $TV$, $TA$, and $VA$. All configurations retain the three intra-modal pathways $T$, $A$, and $V$ and use the same TDL-GDB and KAN fusion mechanism. The None configuration contains no pairwise pathway, the single- and two-pathway configurations retain one or two pairwise pathways, respectively, and Full includes all three. The results on CMU-MOSI and CMU-MOSEI are reported in Table~\ref{tab:interaction_mosi_mosei}, while those on CH-SIMS v2 are reported in Table~\ref{tab:interaction_sims}.

The comparison between \textit{None} and \textit{Full} shows that incorporating all three pairwise pathways consistently improves performance over the intra-modal-only configuration under the current architecture. On CMU-MOSI, the full configuration improves Acc7 and Acc2 by 5.4 and 3.5 percentage points, respectively. The corresponding gains are 5.2 and 1.5 points on CMU-MOSEI, while improvements of 6.3 and 4.2 points are observed for Acc3 and Acc2 on CH-SIMS v2. Consistent reductions in MAE across all three datasets further indicate that the observed overall benefit is not limited to categorical sentiment prediction.

The single- and two-pathway configurations further show that the effectiveness of a particular interaction varies across datasets and evaluation criteria. On CMU-MOSEI, $TA$ achieves the highest Acc7 among the single-pathway variants, whereas $TV$ performs better in Acc2, F1, MAE, and Corr. On CMU-MOSI, the text-centered pathways $TV$ and $TA$ generally provide larger gains than $VA$, while $VA$ becomes comparatively more effective on CH-SIMS v2. These results suggest that the utility of a modality pair is context-dependent rather than fixed across datasets.

Overall, the \textit{Full} configuration provides the most consistent performance across the three benchmarks, although the individual pathways contribute differently under different conditions. Retaining all three pairwise pathways therefore provides the most consistent overall configuration across the three benchmarks for subsequent fusion. This finding motivates the following interpretability analysis, where we further examine how the six pathways and their compressed features participate in the learned KAN decision process.

\subsection{Interpretability Analysis}
\label{subsec:interpretability}

\begin{figure*}[!t]
\centering
\subfloat[CMU-MOSI]{\includegraphics[width=0.32\textwidth]{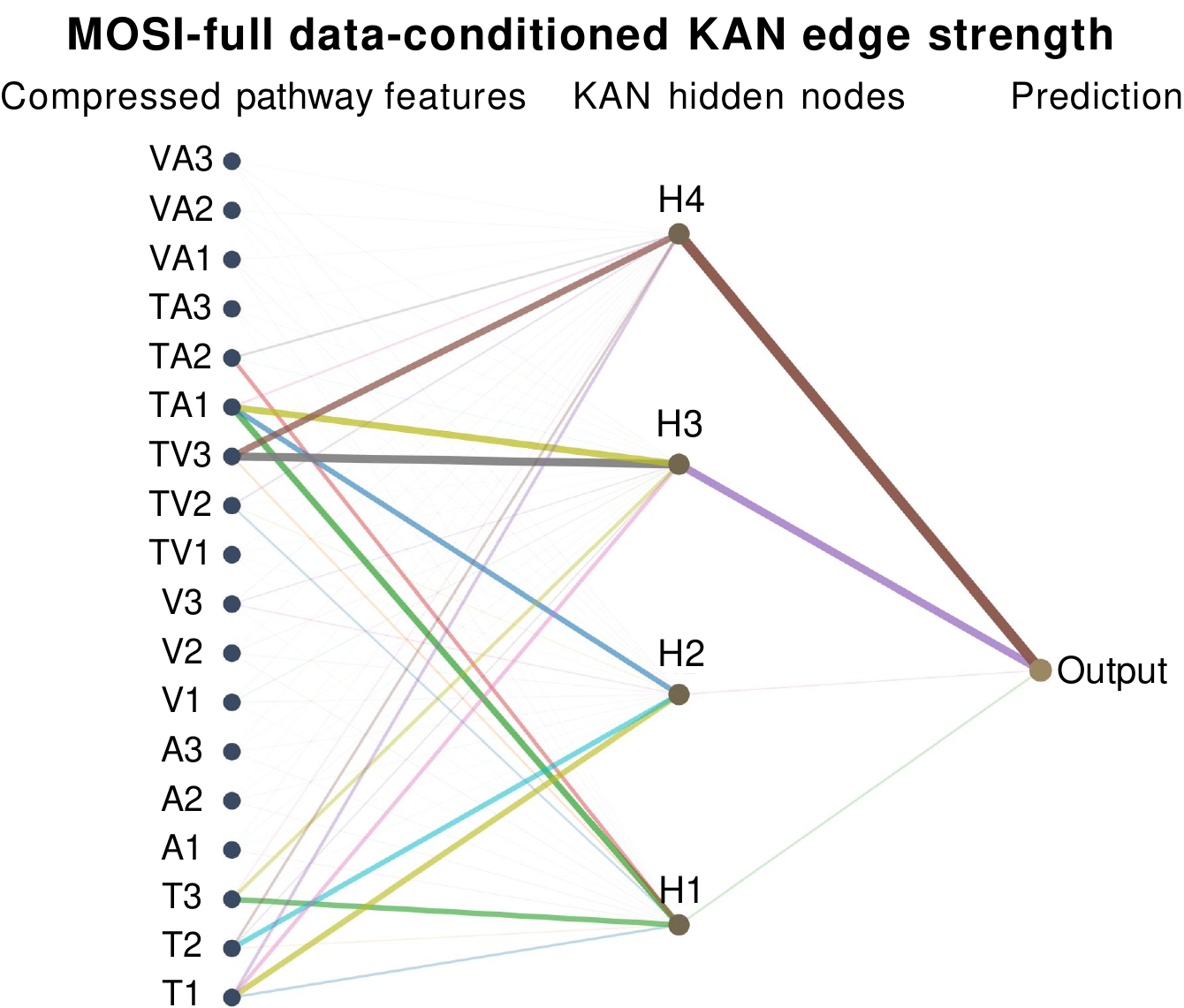}}
\hfill
\subfloat[CMU-MOSEI]{\includegraphics[width=0.32\textwidth]{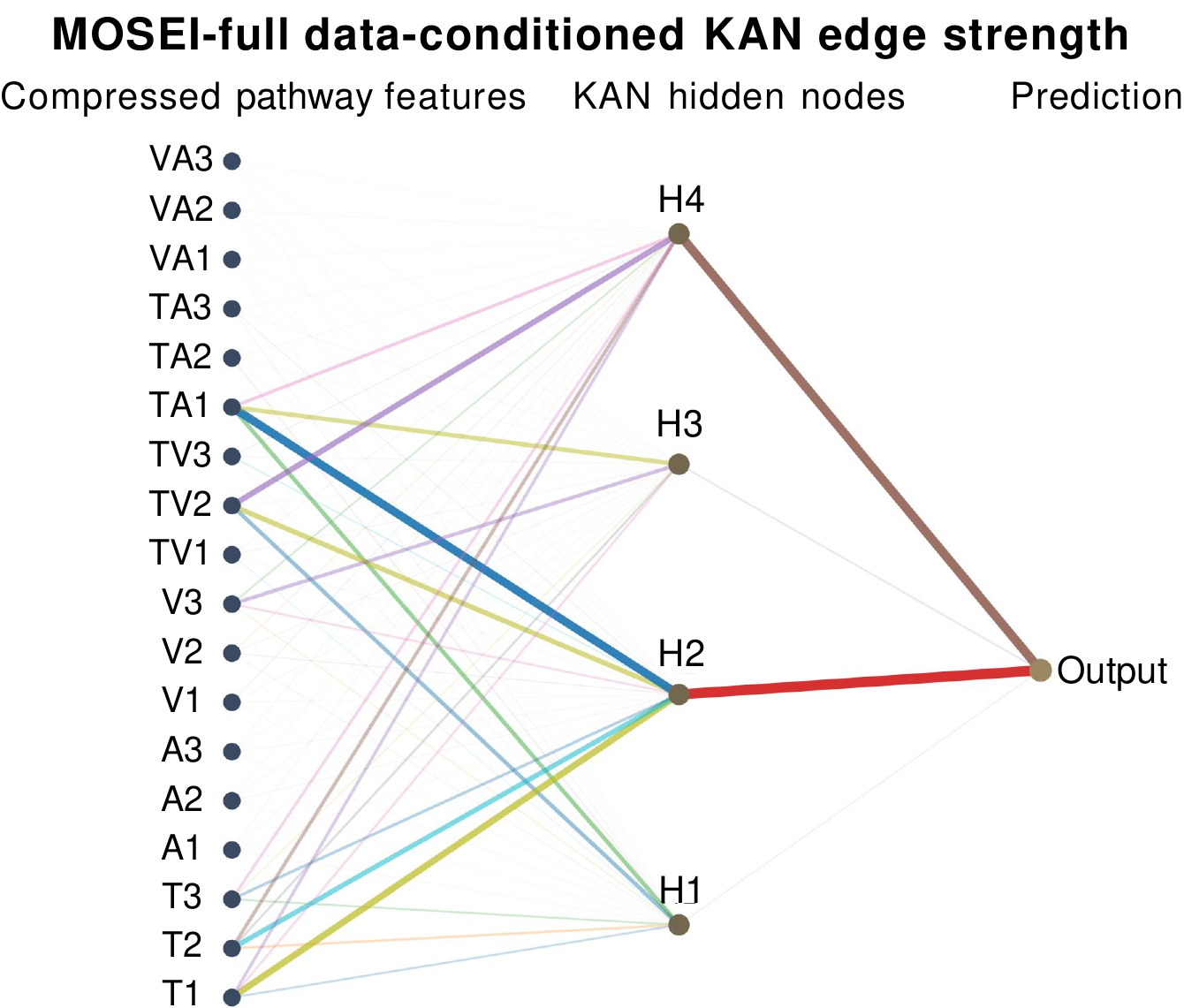}}
\hfill
\subfloat[CH-SIMS v2]{\includegraphics[width=0.32\textwidth]{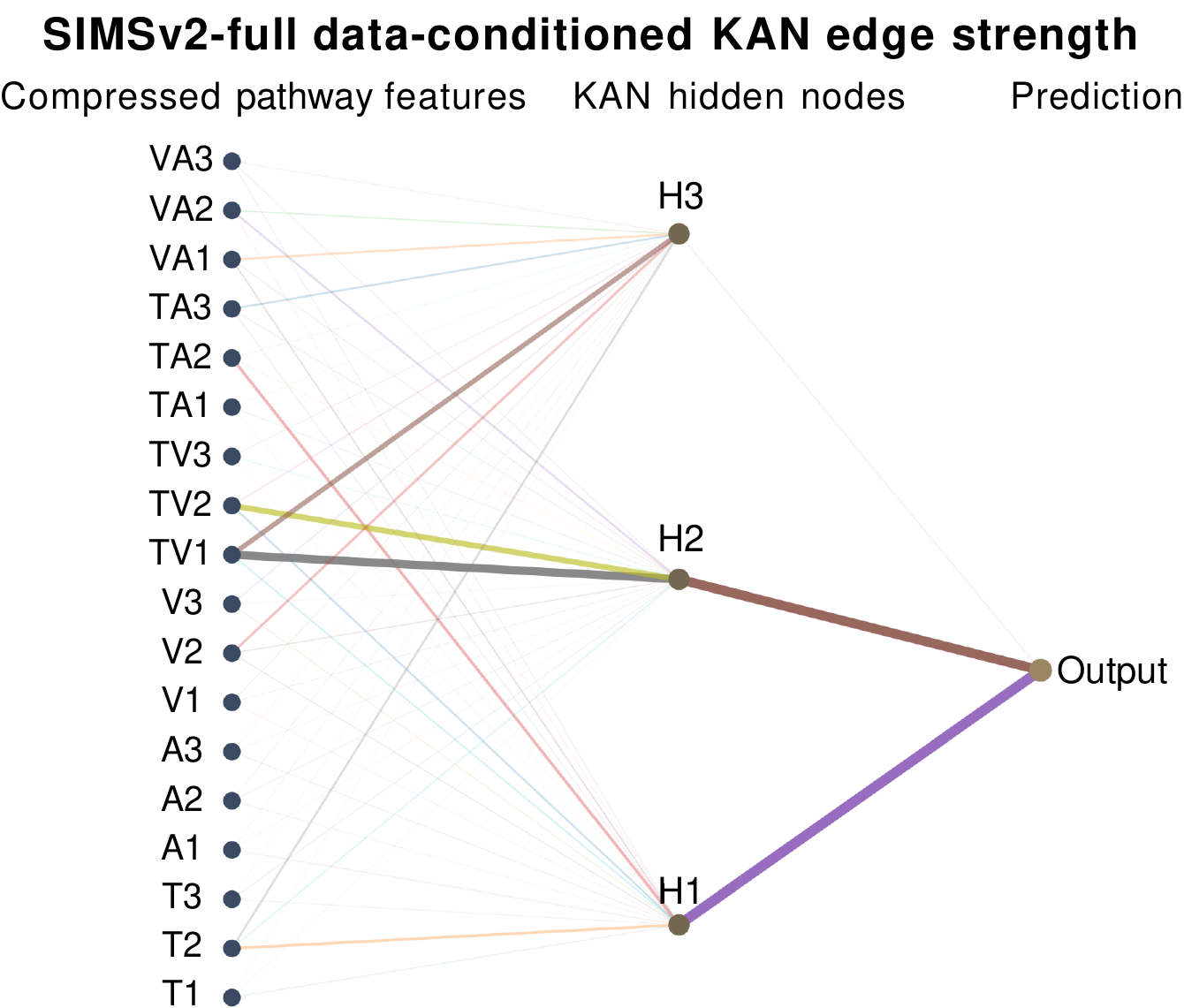}}
\caption{Data-conditioned KAN functional-edge visualization on the three datasets. The left column of each subfigure contains the compressed features from the six intra- and inter-modal pathways, the middle column contains the KAN hidden nodes, and the right node denotes the final sentiment prediction. Edge width and opacity indicate the normalized functional-edge strength.}
\label{fig:edge_strength}
\end{figure*}

The structured six-pathway representation of I$^{2}$-KAN enables the inference process to be examined at three complementary levels: pathway-level predictive sensitivity, compressed-feature predictive sensitivity, and KAN functional-edge behavior. Since each of the six semantic pathways contributes three compressed dimensions to the KAN input, the resulting decision process remains traceable from pathway-level information sources to individual compact features and their nonlinear propagation through KAN.

\subsubsection{KAN Functional-Edge Analysis}

Each KAN connection is associated with a learned univariate transformation rather than a conventional scalar weight. To characterize its data-conditioned behavior, we evaluate the variation of each edge response over the test set and normalize the resulting strength within each KAN layer. Accordingly, thicker and more opaque edges in Fig.~\ref{fig:edge_strength} indicate stronger data-dependent functional responses. Edge colors are used only for visual distinction, and the edge strength should not be interpreted as sentiment polarity or causal contribution.

As shown in Fig.~\ref{fig:edge_strength}, the learned functional responses are concentrated in a subset of nonlinear routes rather than being uniformly distributed across all connections. The input nodes correspond to the compressed dimensions of $T$, $A$, $V$, $TV$, $TA$, and $VA$, while the hidden nodes represent learned aggregation routes without predefined semantic meanings. CMU-MOSI mainly exhibits stronger hidden-to-output responses through $H_3$ and $H_4$, CMU-MOSEI through $H_2$ and $H_4$, and CH-SIMS v2 through $H_1$ and $H_2$. Since the models are trained independently, these hidden-node indices should not be interpreted as semantically aligned across datasets.

The distinct routing patterns are consistent with the different distributions of affective evidence across the three benchmarks. CMU-MOSEI shows a stronger reliance on standalone textual information, whereas CMU-MOSI and CH-SIMS v2 exhibit more interaction-oriented structures, as further confirmed by the pathway-level analysis below. This suggests that KAN does not impose a fixed fusion topology but adapts its nonlinear aggregation routes to dataset-specific multimodal characteristics. The following analysis further quantifies which semantic pathways the final prediction is most sensitive to.

\subsubsection{Pathway-Level Predictive Sensitivity Analysis}

\begin{figure*}[!t]
\centering
\subfloat[CMU-MOSI]{\includegraphics[width=0.32\textwidth]{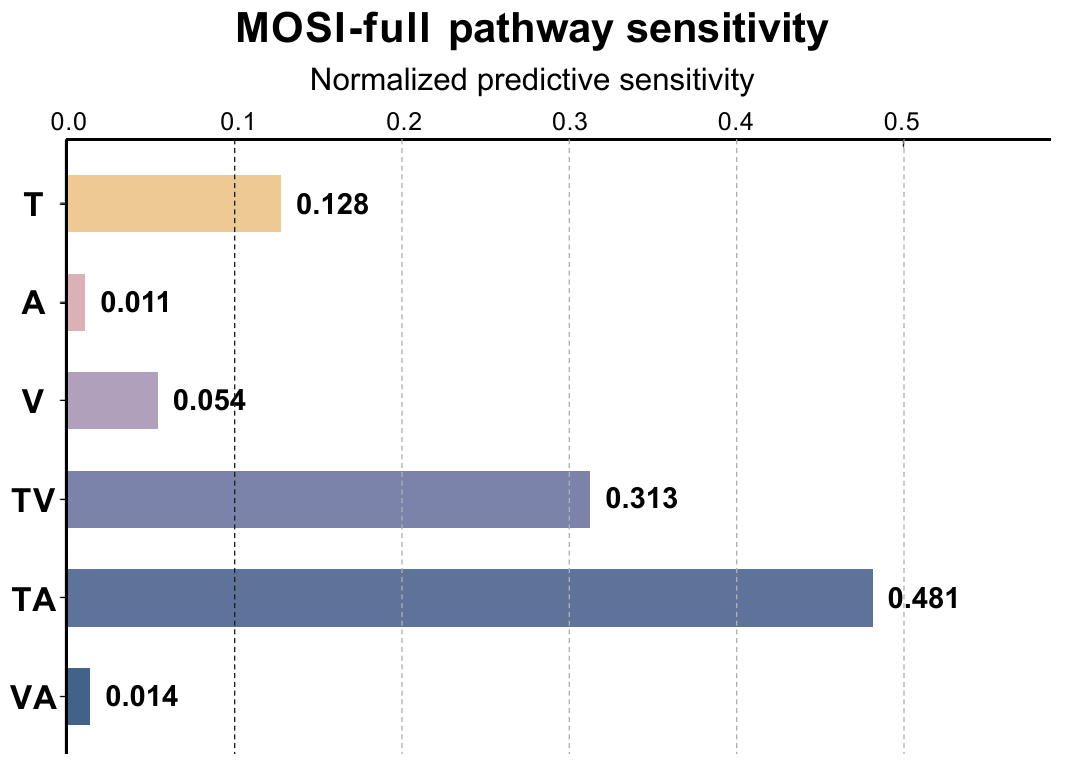}}
\hfil
\subfloat[CMU-MOSEI]{\includegraphics[width=0.32\textwidth]{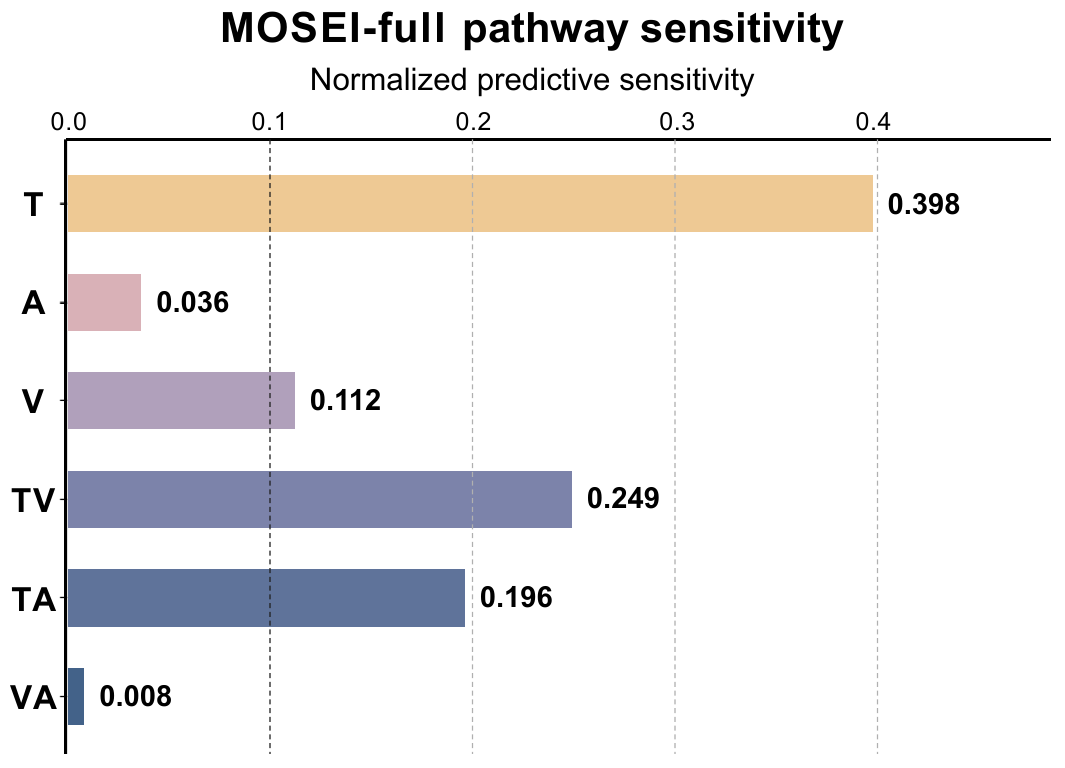}}
\hfil
\subfloat[CH-SIMS v2]{\includegraphics[width=0.32\textwidth]{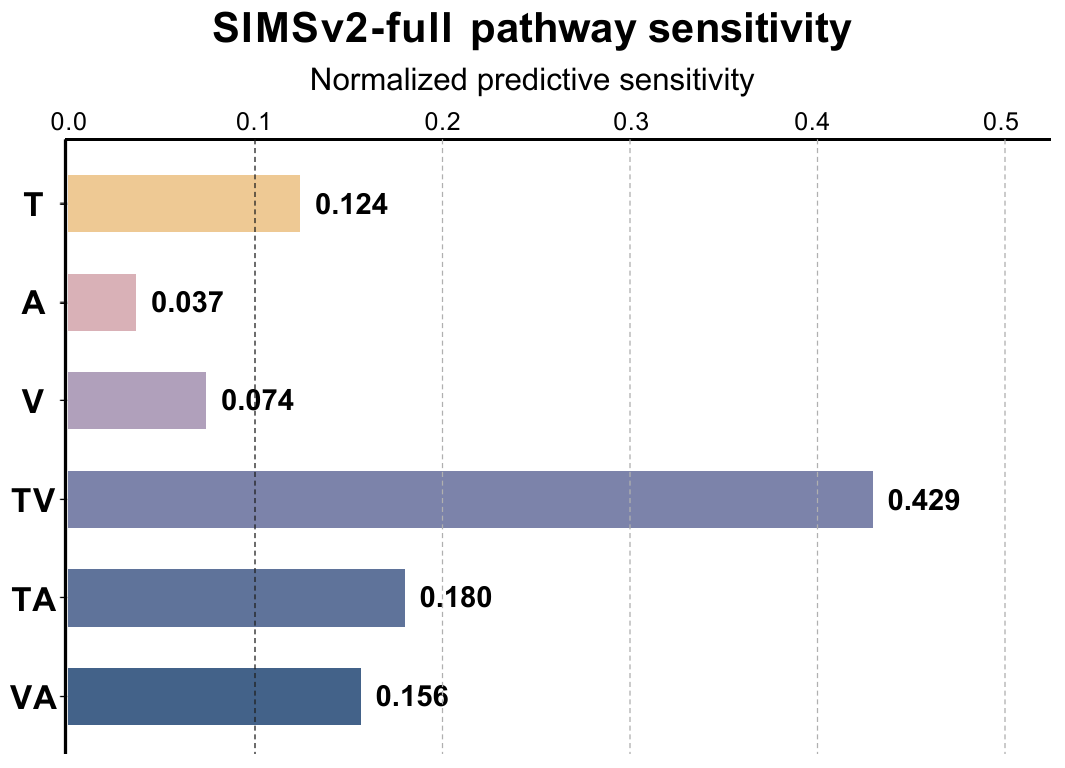}}
\caption{Normalized masking-based predictive sensitivity of the six semantic pathways, obtained by masking all three compressed dimensions belonging to one pathway at a time.}
\label{fig:branch_contribution}
\end{figure*}

To quantify the predictive dependence on each semantic pathway, we mask all three compressed dimensions of one pathway at a time and measure the resulting change in model output. The obtained scores are normalized across the six pathways for comparison, as shown in Fig.~\ref{fig:branch_contribution}.

As shown in Fig.~\ref{fig:branch_contribution}, pairwise interaction pathways constitute a substantial part of the learned decision structure across all three datasets. A particularly consistent pattern appears for the acoustic modality: although the standalone $A$ pathway exhibits low predictive sensitivity, the audio-related interaction pathways $TA$ and $VA$ can exhibit considerably higher predictive-sensitivity scores. This observation across different datasets supports our central argument that low standalone predictive utility does not necessarily imply low interaction-conditioned predictive utility.

The dominant pathways also differ across datasets. CMU-MOSI is strongly interaction-oriented, with $TA$ and $TV$ exhibiting the highest predictive-sensitivity scores, whereas CMU-MOSEI shows markedly higher sensitivity to the standalone textual pathway, with $T$ reaching a normalized score of $0.398$. This is consistent with the large-scale opinion-video setting of CMU-MOSEI, where explicit linguistic sentiment often provides stable direct evidence \cite{zadeh2018multimodal}. In contrast, CH-SIMS v2 exhibits higher predictive sensitivity to $TV$ and $VA$, which is consistent with its richer nonverbal affective cues and greater reliance on cross-modal contextual information \cite{liu2022make}.

Despite these dataset-specific differences, text remains an important semantic anchor. These results suggest that textual information often exerts its influence not only through the standalone $T$ pathway, but also through explicit interactions with acoustic and visual evidence.

\subsubsection{Compressed-Feature Predictive Sensitivity Analysis}

\begin{figure*}[!t]
\centering
\subfloat[CMU-MOSI]{\includegraphics[width=0.32\textwidth]{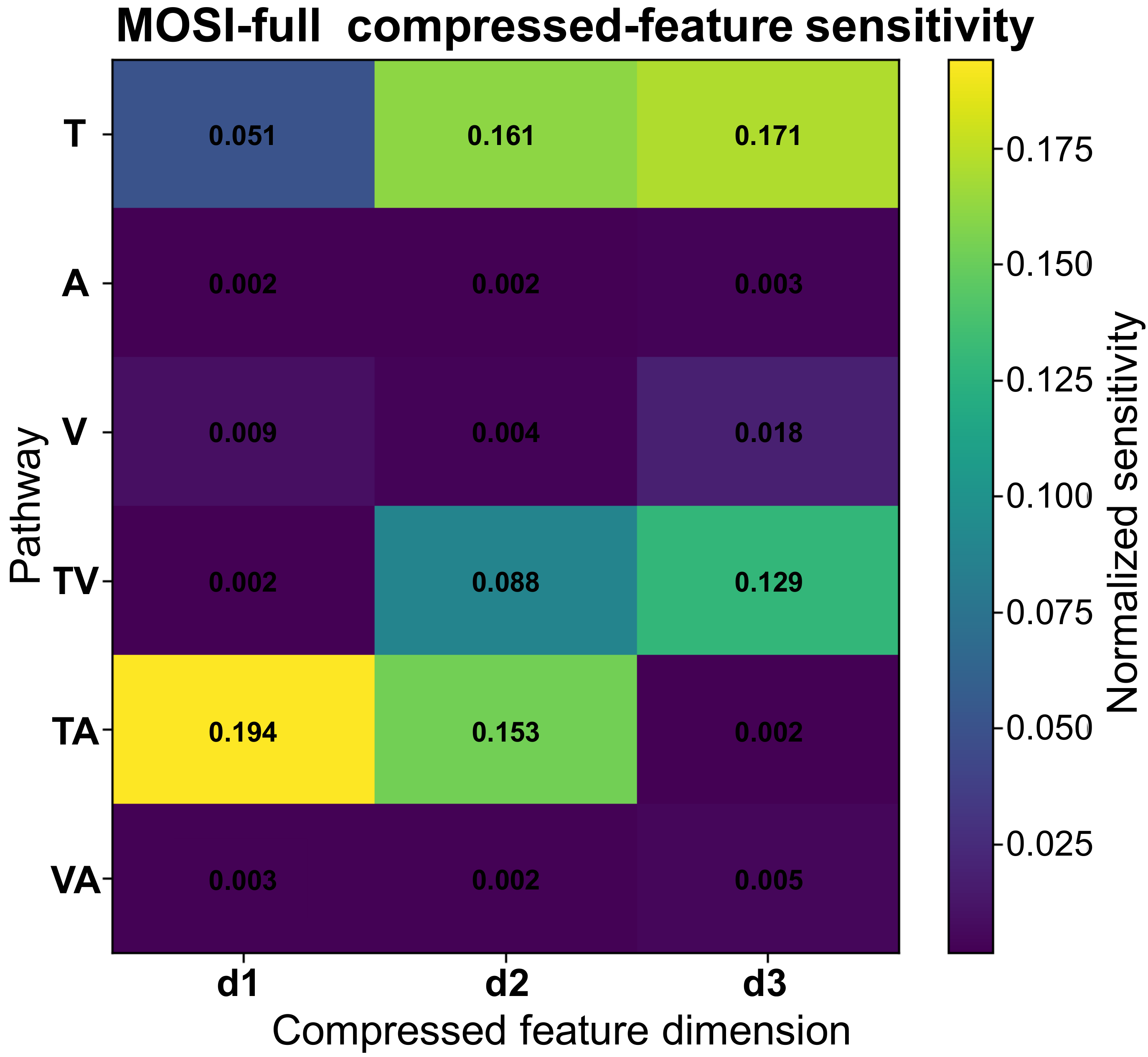}}
\hfill
\subfloat[CMU-MOSEI]{\includegraphics[width=0.32\textwidth]{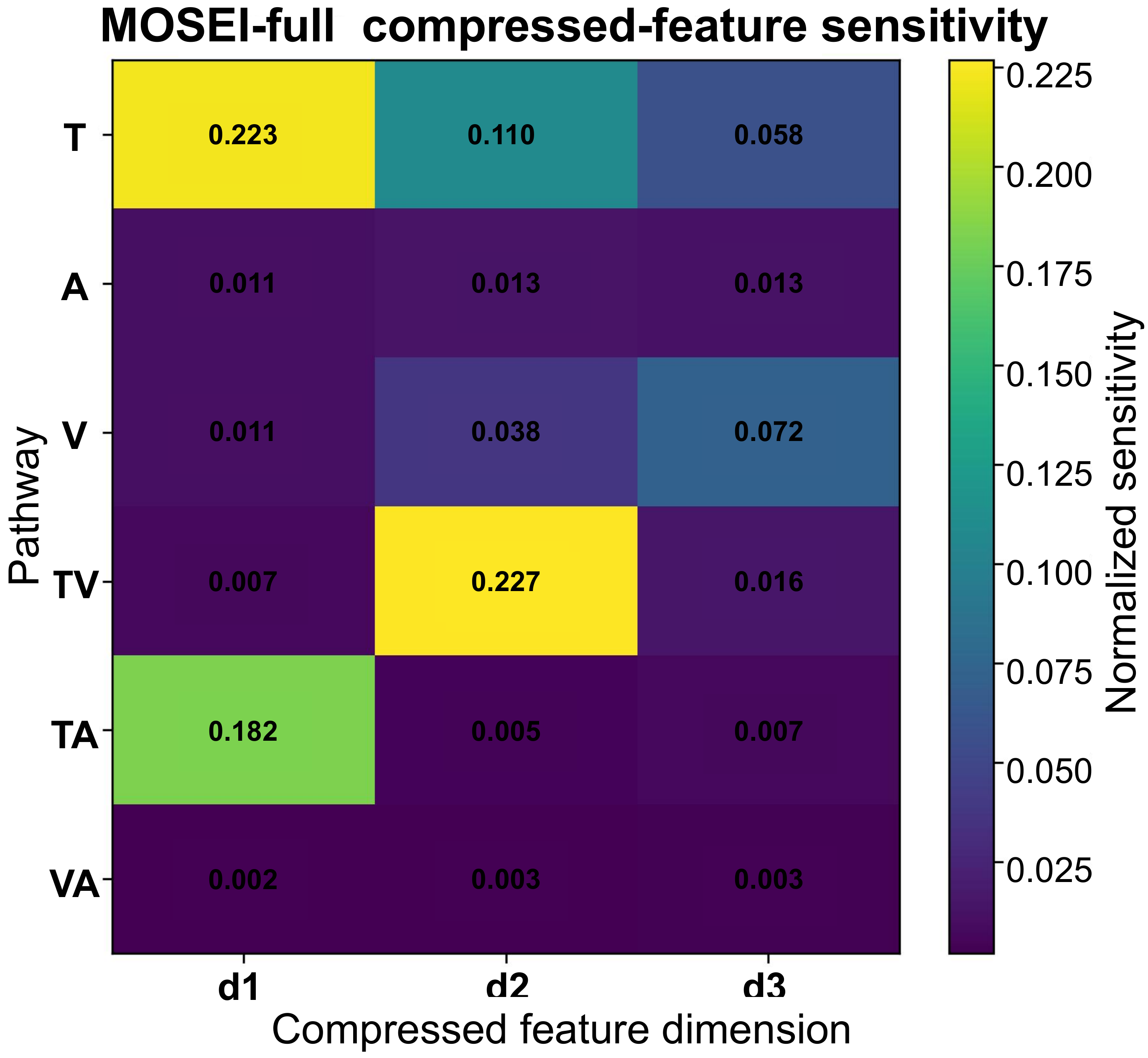}}
\hfill
\subfloat[CH-SIMS v2]{\includegraphics[width=0.32\textwidth]{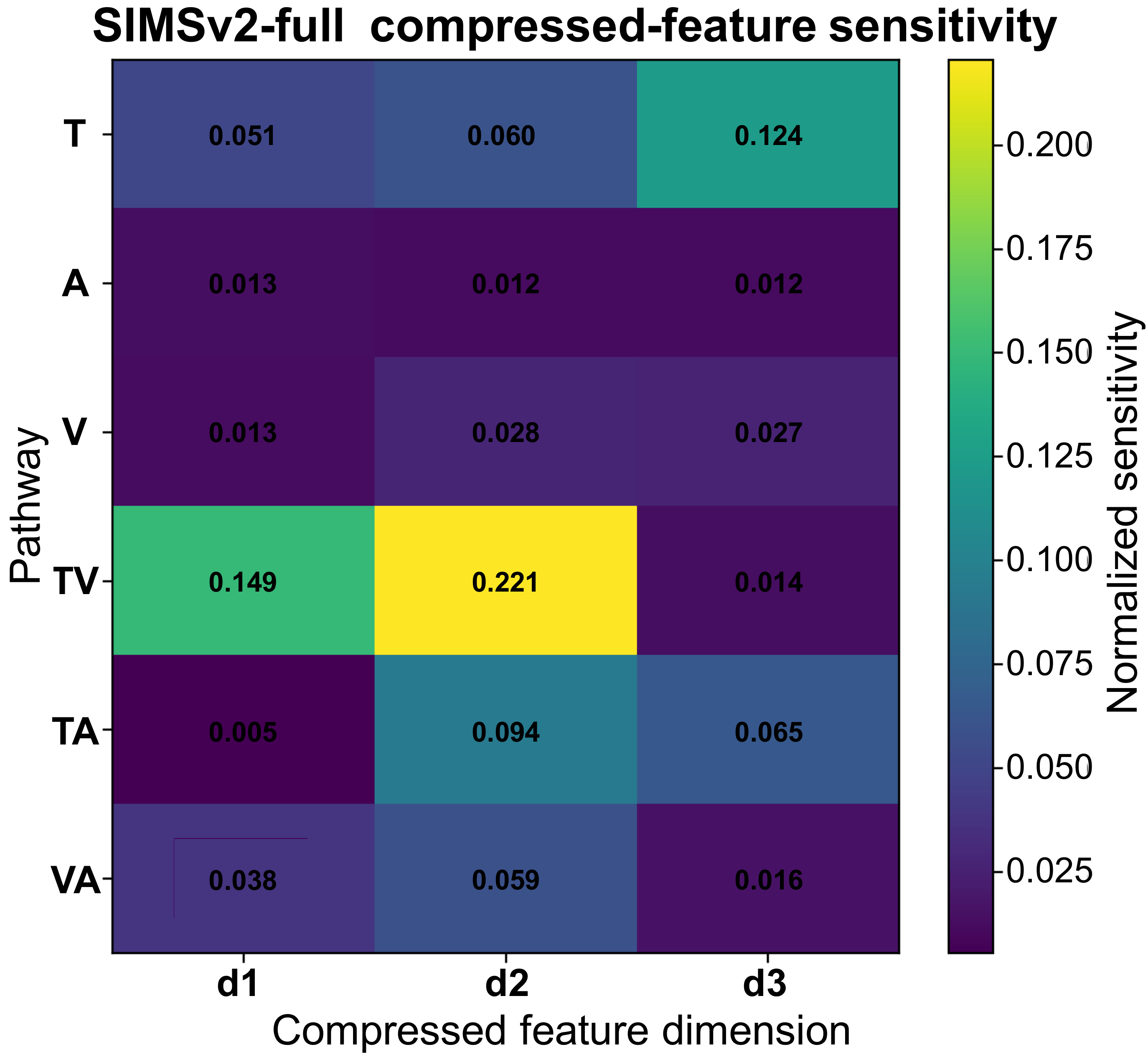}}
\caption{Normalized masking-based predictive sensitivity of individual compressed dimensions within each pathway.}
\label{fig:feature_contribution}
\end{figure*}

To further examine how predictive information is distributed within each semantic pathway, we mask one compressed dimension at a time and measure the resulting change in model output. The normalized results are visualized in Fig.~\ref{fig:feature_contribution}.

As shown in Fig.~\ref{fig:feature_contribution}, the three compressed dimensions within the same pathway exhibit clearly different predictive-sensitivity patterns. The dimensions with higher sensitivity also vary across semantic pathways and datasets rather than following a fixed pattern, suggesting that the bottleneck captures compact features with differentiated predictive roles rather than uniformly redundant representations.

The variation suggests that feature specialization depends jointly on the semantic source of a pathway and the underlying data distribution. For example, text-related dimensions are more prominent on CMU-MOSEI, whereas interaction-related dimensions are more prominent on CMU-MOSI and CH-SIMS v2, consistent with the pathway-level findings above. Since these compact dimensions are learned independently for each dataset, their indices should not be interpreted as carrying fixed semantic meanings across models.

Together with the pathway-level analysis, these results show that the information retained by I$^{2}$-KAN is structured not only across semantic pathways but also within each pathway, providing a finer-grained view of how the final prediction depends on compact multimodal features.

\subsection{Ablation Study}
\label{subsec:component_ablation}

\begin{table}[!t]
\centering
\caption{Leave-one-component-out ablation results on CMU-MOSEI.}
\label{tab:component_ablation}
\renewcommand{\arraystretch}{1.10}
\setlength{\tabcolsep}{4.0pt}
\resizebox{\columnwidth}{!}{
\begin{tabular}{l|ccccc}
\hline
\tblhead{Configuration} & \tblhead{Acc7$\uparrow$} & \tblhead{Acc2$\uparrow$} & \tblhead{F1$\uparrow$} & \tblhead{MAE$\downarrow$} & \tblhead{Corr$\uparrow$} \\
\hline
w/o Pairwise Interactions & 49.2 & 86.7 & 86.6 & 0.578 & 0.753 \\
w/o TDL-GDB & 53.8 & 85.6 & 85.6 & 0.525 & 0.778 \\
w/o KAN & 53.0 & 86.7 & 86.6 & 0.524 & 0.780 \\
I$^{2}$-KAN & \textbf{54.4} & \textbf{88.2} & \textbf{88.1} & \textbf{0.515} & \textbf{0.798} \\
\hline
\end{tabular}}
\end{table}

We conduct a leave-one-component-out study on CMU-MOSEI, the largest benchmark considered in this work, to examine the roles of pairwise interaction pathways, TDL-GDB, and KAN fusion. Each variant modifies only one major component while keeping the remaining settings unchanged. The \textit{w/o Pairwise Interactions} variant removes $TV$, $TA$, and $VA$ while retaining the three intra-modal pathways with TDL-GDB and KAN. The \textit{w/o TDL-GDB} variant retains all six pathways and KAN but replaces TDL-GDB with a standard low-dimensional bottleneck of the same output dimensionality. The \textit{w/o KAN} variant retains the six pathways and TDL-GDB but replaces KAN with an MLP of comparable hidden width.

The corresponding ablation results are reported in Table~\ref{tab:component_ablation}. As shown in Table~\ref{tab:component_ablation}, removing the pairwise interaction pathways leads to a clear performance degradation across all metrics, indicating that incorporating $TV$, $TA$, and $VA$ is beneficial beyond the intra-modal-only configuration under the current architecture. This result is consistent with the interaction-combination study and the pathway-level interpretation presented above.

Replacing TDL-GDB with a conventional bottleneck also degrades performance, showing that simple dimensionality reduction is insufficient for learning compact representations from heterogeneous intra-modal and interaction pathways. This comparison supports the effectiveness of task-decoupled denoising and gated information selection in TDL-GDB.

Finally, replacing KAN with a comparable-width MLP results in lower performance while retaining the same six pathways and TDL-GDB representations. This suggests that the gains of I$^{2}$-KAN do not arise solely from introducing additional interaction pathways, and that KAN provides an effective nonlinear fusion mechanism for the structured pathway representation while retaining the functional interpretability analyzed in Section~\ref{subsec:interpretability}.

\section{Conclusion}
\label{sec:conclusion}

In this paper, we proposed I$^{2}$-KAN, an interpretable multimodal sentiment analysis framework that explicitly organizes intra-modal and pairwise inter-modal information within a structured multi-pathway representation. TDL-GDB learns compact and task-adaptive representations for heterogeneous pathways, while structured KAN fusion retains explicit pathway associations during nonlinear sentiment prediction. Experiments on CMU-MOSI, CMU-MOSEI, and CH-SIMS v2 demonstrate the effectiveness of the proposed framework, with interaction-pathway and component-wise ablations further supporting the proposed architectural design. More importantly, the multi-level analyses consistently show that standalone predictive utility does not necessarily reflect interaction-conditioned predictive utility. These findings highlight the interaction-dependent nature of multimodal information and suggest that modality utility should be considered from both standalone and interaction-conditioned perspectives. By maintaining traceable pathway associations throughout representation refinement and fusion, I$^{2}$-KAN provides a finer-grained view of multimodal sentiment reasoning.

\bibliographystyle{IEEEtran}
\bibliography{references}

\end{document}